\documentclass[10pt,twocolumn,letterpaper]{article}

\usepackage[pagenumbers]{iccv} %

\usepackage{multirow}
\usepackage{array} %
\usepackage{ragged2e} %
\usepackage{graphicx}
\usepackage{enumitem}
\usepackage{amsmath}
\usepackage{caption}
\usepackage{xcolor}
\usepackage{afterpage} %

\newcommand{\OURS}{ExCap3D}

\newcommand{\DatasetName}{\OURS{} Dataset}
\newcommand{\DatasetNumScenes}{947}
\newcommand{\DatasetNumObjects}{34k}
\newcommand{\DatasetNumTotalCaptions}{190k}

\newcommand{\ObjectCiderImprovementPercentage}{17}
\newcommand{\PartCiderImprovementPercentage}{124}

\definecolor{iccvblue}{rgb}{0.21,0.49,0.74}
\usepackage[pagebackref,breaklinks,colorlinks,allcolors=iccvblue]{hyperref}

\begin{document}
\title{\OURS: Expressive 3D Scene Understanding via Object Captioning \\ with Varying Detail}

\author{Chandan Yeshwanth\qquad David Rozenberszki \qquad Angela Dai\\
\\
Technical University of Munich
} 

\twocolumn[{
\renewcommand\twocolumn[1][]{#1}
\maketitle
    \begin{center}
        \centering
        \captionsetup{type=figure}
        \includegraphics[width=\textwidth]{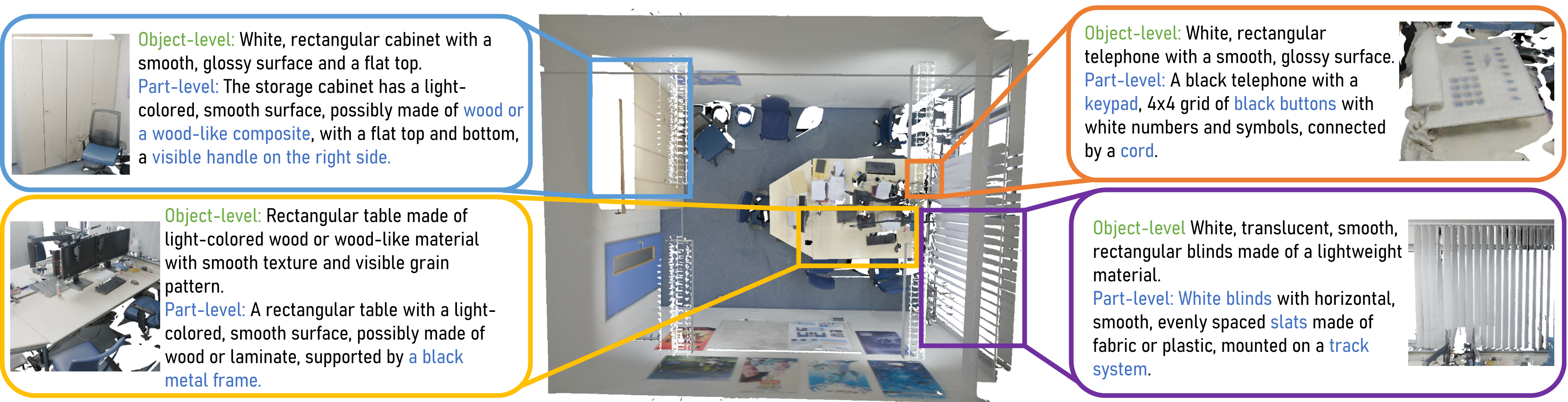}
        \captionof{figure}{\OURS{} produces descriptions at multiple level of details for objects in an input 3D scan. For each detected object, we generate both object- and part-level of detail captions. The generated captions are consistent across the two levels and contain rich details.}
        \label{fig:teaser}
    \end{center}
}]

\begin{abstract}

Generating text descriptions of objects in 3D indoor scenes is an important building block of embodied understanding. 
Existing methods describe objects at a single level of detail and do not capture fine-grained details of the parts of objects.
In order to produce varying levels of detail capturing both coarse object-level information and detailed part-level descriptions,  we propose the task of expressive 3D captioning. Given an input 3D scene, the task is to describe objects at multiple levels of detail: a high-level object description, and a low-level description of the properties of its parts.
To produce such captions, we present \OURS{}, an expressive 3D captioning model which takes as input a 3D scan, and for each detected object in the scan, generates a fine-grained collective description of the parts of the object, along with an object-level description conditioned on the part-level description.
We design \OURS{} to encourage consistency between the multiple levels of  descriptions.
To enable this task, we generated the \DatasetName{} by leveraging a visual-language model (VLM) for multi-view captioning. The \DatasetName{} contains captions on the ScanNet++ dataset with varying levels of detail,
comprising \DatasetNumTotalCaptions{} text descriptions of \DatasetNumObjects{} 3D objects in \DatasetNumScenes{} indoor scenes.
Our experiments show that the object- and part-level details generated by \OURS{} are more expressive than those produced by state-of-the-art methods, with a CIDEr score improvement of \ObjectCiderImprovementPercentage{}\% and \PartCiderImprovementPercentage{}\% for object- and part-level details respectively. Our code, dataset and models will be made publicly available.
\end{abstract}
    
\vspace{-0.4cm}
\section{Introduction}
\label{sec:intro}
\vspace{-0.2cm}

3D indoor scene understanding is central to applications such as augmented reality (AR), virtual reality (VR), and robotics. While detecting 3D objects and their semantics is important for these applications, natural language descriptions of objects enable more wide-ranging possibilities -- they can encode more complex information efficiently and, importantly, enable more natural interactions between embodied agents as well as humans and the scene. To produce a representation that facilities a diverse array of applications, such language descriptions must also encompass varying levels of detail: object-level descriptions for coarse identification in the scene, and part-level details for precise manipulation and fine-grained understanding.
For instance, a robotic assistant would only need a coarse, high-level description of the objects in a room (e.g., `a recliner chair') in order to navigate effectively in it, but it would need fine-grained descriptions of the objects (e.g., `recliner chair with a high padded back, wooden armrests, a soft cushion and an adjustable footrest') to provide assistive AI descriptions for accessibility aid or to describe the scene to others.

\begin{figure}
\includegraphics[width=\linewidth]{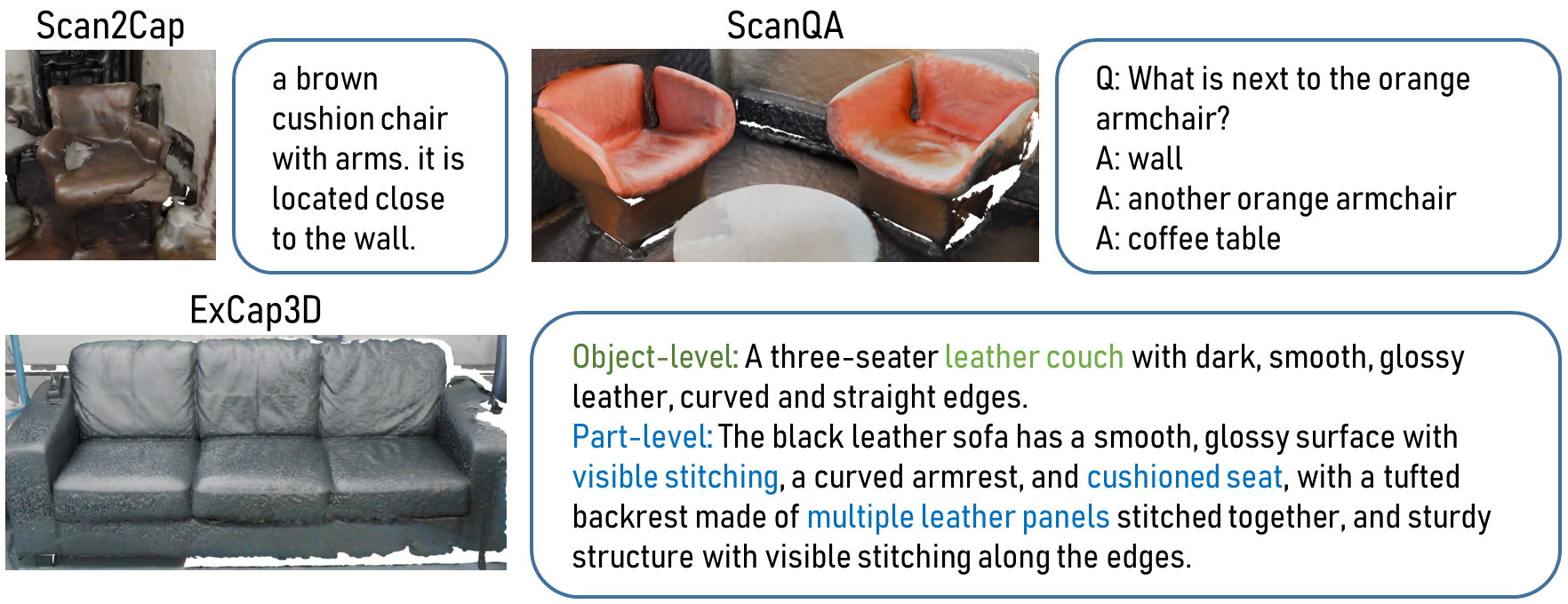}
\caption{Descriptions of objects in existing datasets such as Scan2Cap and ScanQA largely contain relations between objects, and limited local details at a single level. In contrast, the \DatasetName{} describes both the object as a whole and as a sum of its parts.
}
\label{fig:datasets-comparison-teaser}
\end{figure}

Current 3D captioning models have made significant progress in describing objects in indoor scenes by detecting objects and generating their individual descriptions at a single level of detail \cite{chen2023vote2cap,chen2021scan2cap,kim2024seeitall,zhu20233dvista, zhang2024chatscene, chen2024ll3da,zhu2024unifyingpq3d}.
Such captions can effectively describe coarse object information in the global context of the scene, but preclude detailed perception of objects.
This is because they contain limited detail of the objects themselves and fail to describe highly localized properties of their parts, such as appearance, material, and function. In order to enable more complex perception of a scene requiring these details, we propose to describe objects at multiple levels:
an object-level of detail that captures its most prominent aspects (semantic class; possibly size or overall appearance), and a finer-grained part-level of detail that describes the details of elements composing the object (different materials, color, and functions). These two descriptions are complementary to each other and enable different applications depending on the level of detail required. Examples of such multilevel descriptions are shown in Fig. \ref{fig:teaser} and Fig. \ref{fig:datasets-comparison-teaser}.

To generate such multilevel captions in 3D scans, we introduce \OURS{}, a 3D object captioning model that produces expressive captions at both object- and part-levels of detail using two captioning heads.
Examples of the captions generated by our method are shown in Fig. \ref{fig:teaser}. The generated multilevel descriptions are complementary to each other and describe different aspects of the objects, while maintaining a few shared elements for consistency and coherence.
The part-level caption generated by \OURS{} collectively describes the different parts of an object. Knowing the descriptions of the parts can enable our model to generate more consistent captions describing the object as a whole. Hence, \OURS{} first produces the part-level details and then uses these to inform the object-level captioner. To further reduce inconsistencies between the two levels of detail in terms of the object semantics and text content, \OURS{} employs additional consistency losses during training.

To enable this multilevel captioning task, we provide the \DatasetName{}, which contains \DatasetNumTotalCaptions{} object and part-level natural language captions of \DatasetNumObjects{} 3D objects from the \DatasetNumScenes{} scenes of the ScanNet++ dataset. 
We automatically generate object captions for training, using a pretrained vision-language model (VLM) on high-resolution DSLR images along with ground truth semantic information from the ScanNet++ dataset~\cite{yeshwanthliu2023scannetpp}.
We employ multi-view aggregation to robustly capture key object- and part-level details. A comparison of the multilevel captions from the \DatasetName{} with the single-level descriptions from the existing datasets Scan2Cap \cite{chen2021scan2cap} and ScanQA \cite{azuma2022scanqa} is shown in Fig. \ref{fig:datasets-comparison-teaser}. The \DatasetName{} describes objects with significantly more detail, while providing two complementary multilevel descriptions of each object.

We show that \OURS{} produces richer and more consistent object- and part-level captions compared to the state-of-the-art captioning methods, originally designed to generate only a single level description. 
To summarize, our main contributions are:
\begin{itemize}
    \item an expressive 3D captioning model that generates captions of objects in 3D indoor scenes at multiple levels of detail: fine-grained part-level details and object-level details
    \item modeling object-level detail as a sum of its parts' unique appearance and functional attributes by generating object-level captions conditioned on part-level descriptions, using the cross-attention mechanism of the object captioner and the language model's intermediate hidden states
    \item regularization of our joint 3D captioning approach through semantic and textual-similarity consistency losses that maintain consistency of the object-level and part-level predicted captions
    \item the \DatasetName{}: a large-scale, precise and challenging dataset of \DatasetNumTotalCaptions{} descriptive object- and part-level captions of \DatasetNumObjects{} objects in 3D indoor scenes built on the ScanNet++ dataset with fine-grained texture, material, and functional descriptions
\end{itemize}

\section{Related Work}
\label{sec:related-work}

\subsection{2D Image Captioning}
2D image captioning has been extensively studied in computer vision. DenseCap~\cite{johnson2016densecap} popularized the task of dense captioning of images, jointly localizing objects, and generating their short descriptions. 
More recently, COCONut-PanCap~\cite{deng2025coconut} and Pix2Cap-COCO~\cite{you2025pix2cap} combined the tasks of fine-grained instance segmentation and detailed captioning.
FlexCap~\cite{dwibedi2024flexcap} and FineCaption~\cite{hua2024finecaption}  can produce descriptions at varying levels of detail given the image and a localized object as input. In contrast, we combine the tasks of detection and multilevel captioning and describe detected objects and their parts.

\subsection{3D Scene Captioning}

Scan2Cap~\cite{chen2021scan2cap} introduced the task of 3D dense captioning and a method for context-aware captioning of 3D objects, operating on ScanNet~\cite{dai2017scannet} scenes. Since then, several methods have been developed to effectively solve this task by extracting object features and modeling their relations with nearby objects \cite{yuan2022x, chen2023vote2cap, jiao2022more, wang2022spatiality, chen2022d3net, cai20223djcg, kim2024see}. These methods can accurately describe objects at the scene-level through relations with neighboring objects.
Following the trends of utilizing foundation models in 3D reasoning, Cap3D~\cite{luo2023scalablecap3d} and follow-up works \cite{luo2024view, ge2024visual} describe isolated 3D objects through multi-view aggregation of VLM-generated captions. 
More recent works have jointly trained on multiple tasks such as 3D grounding, question answering, and dense captioning \cite{chen2023unit3d, zhu2024unifyingpq3d, hong20233d, huang2024chat, zhu2024llava, wang2023chat,zhu20233d,chen2024ll3da}, while leveraging advances in large language models (LLMs) to produce more natural and accurate object captions. 
However, these works focus on high-level, proximity-based relations between objects at the scene-level, while producing limited object-level details and no part-level details. 
We tackle a similarly inspired problem, and leverage the high-fidelity geometry and RGB captures from ScanNet++~\cite{yeshwanthliu2023scannetpp} to enable complex, multilevel joint captioning. Moreover, we demonstrate that our joint captioning approach improves both levels of detail.

\subsection{Part Understanding of 3D Objects in Scenes}
Several methods have focused on part-level understanding of 3D scenes by only relying on explicit geometry \cite{bokhovkin2021towards, bokhovkin2023neural}, images and radiance fields \cite{bhalgat2024n2f2, kim2024garfield} or by using both modalities and mesh-based oversegmentation \cite{takmaz2025search3d}. 
A related method for 3D shapes rather than scenes was proposed by Chen et. al.~\cite{chen2024reasoning3d}, which produces part-level segmentation maps and text descriptions of individual 3D objects at a single, fine-grained level of detail. 
This produces promising detailed captions, but is designed for isolated objects rather than complex 3D scenes. 
In contrast, we jointly detect and caption objects in large 3D scenes at multiple levels of detail, ensuring consistency and enabling greater expressivity for downstream tasks.

\section{Method}
\label{sec:method}

\paragraph{Problem Formulation}
The task of expressive captioning is to detect the objects in a 3D scene and describe each object at different levels of detail: object- and part-levels. 
Given an input 3D scene in the form of a 3D mesh  $M = (\mathcal{V}, \mathcal{F})$ with vertices $\mathcal{V} \in \mathbb{R}^{N_{v} \times 3}$ and triangular faces $f_i \in \mathcal{F}$ connecting the vertices, we aim to detect the 3D object instances $o_i = (b_i, s_i, c_{\mathrm{obj},i}, c_{\mathrm{part},i})$ in the scene, characterized by the objects' bounding boxes $b_i$, semantic class $s_i$, and $c_{\mathrm{obj},i}$ and $c_{\mathrm{part},i}$ describing the object in natural language at object- and part-levels of detail, respectively.

\paragraph{Overview of \OURS{}}
First, \OURS{} performs instance segmentation and instance-aware query generation to produce queries containing rich per-object information. Next, the object- and part-level captioning models with shared information generate two separate text captions, conditioned on individual instance-aware queries and the corresponding features from the instance segmentation backbone. 
Finally, semantic and whole-text consistency losses are applied to the hidden states of the captioning models to regularize the consistency of the predicted captions.

\begin{figure*}
\centering
\includegraphics[width=0.9\linewidth]{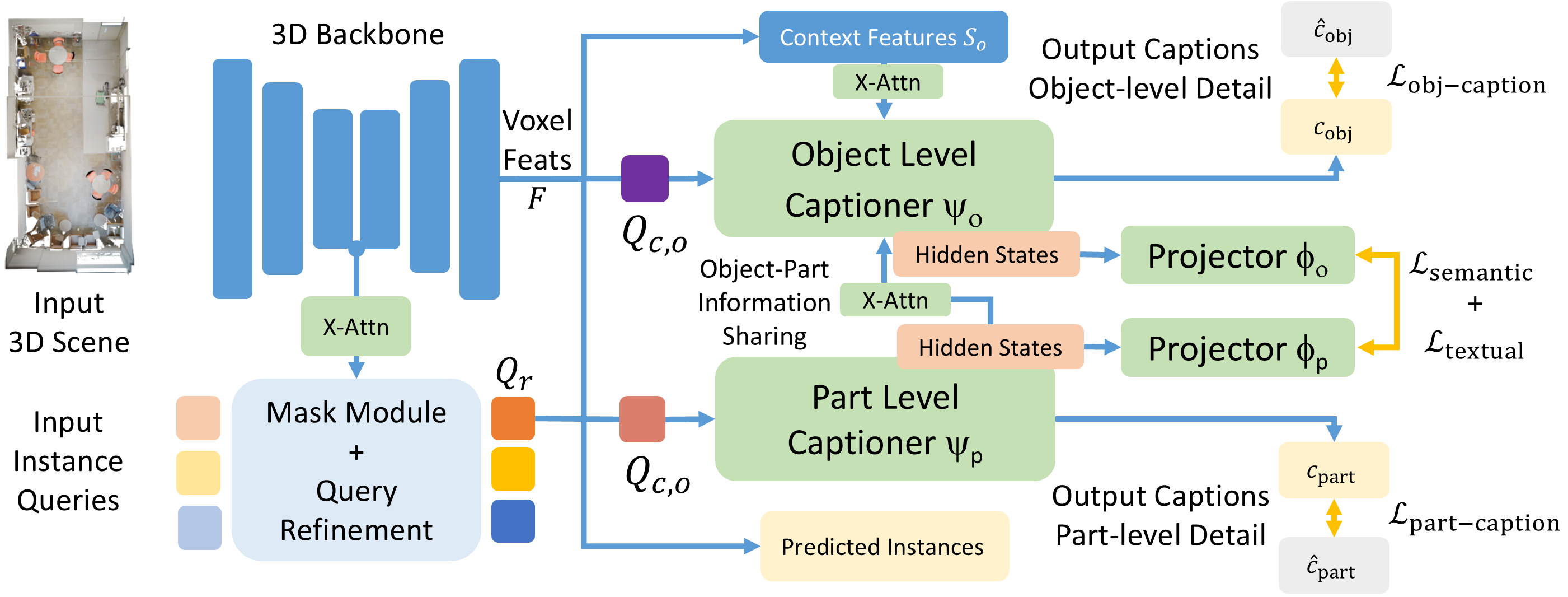}
\caption{Overview of our \OURS{} captioning method. We predict 3D instances in the input 3D scene using Mask3D \cite{schult2023mask3d}, then predict object-level and part-levels of detail using two separate captioning heads. The object-level captions are further constrained by the part-level captioner's hidden states. Semantic- and textual consistency losses are applied to ensure the overall consistency of both predicted captions.}
\label{fig:method-overview}
\end{figure*}

\subsection{3D Instance Segmentation}
Given the mesh of the 3D scene $M$ as input, the instance segmentation model produces a fixed number of refined instance-aware queries $Q_r$ corresponding to the $N_q$ objects detected in the scene, and their binary instance masks on the voxelized 3D input $I \in \{0,1\}^{N_q \times N_{\mathrm{vox}}}$. 
The instance queries are a compact representation of the 3D features of the objects as well as their semantics. 
The instance segmentation model also produces dense features $F \in \mathbb{R}^{N_{\mathrm{vox}} \times D}$ on the voxels.  
The refined and instance-aware queries $Q_r$ and 3D voxel features $F$ are used to inform the per-instance captioning in the next step.

The input mesh is first voxelized into a voxel grid $V \in \mathbb{R}^{N_{\mathrm{vox}} \times 6}$ containing the voxel coordinates and RGB colors.
We use the Mask3D \cite{schult2023mask3d} transformer-based model for instance segmentation. 
It consists of a 3D sparse convolutional UNet backbone to encode the voxelized 3D scene, which outputs features $F \in \mathbb{R}^{N_{\mathrm{vox}} \times D}$ over the input voxels. 
A parallel mask module takes as input query vectors $Q \in \mathbb{R}^{N_q \times D_q}$ 
 denoting the $N_q$ instances in the scene. 
 The queries are refined iteratively through several transformer encoder layers, with cross-attention to the 3D backbone features. The output instance-aware query vectors contain rich representations of the object instances in the 3D scene and are used to predict the instance masks $I \in \{0,1\}^{N_q \times N_{\mathrm{vox}}}$ and semantic class of each object.

\subsection{Joint Captioning with Information Sharing}
\OURS{} contains two separate captioners for the object- and part-levels of details. These are transformer-based language models $\Psi_{\mathrm{obj}}$ and $\Psi_{\mathrm{part}}$ that produce the object-level caption $c_{\mathrm{obj},o}$, part-level caption $c_{\mathrm{part},o}$. 
They are conditioned on the instance-aware queries $Q_r$ and the dense voxel features $F$, and are trained for next-token prediction. The language model captioners take as input a sequence of previously predicted tokens  $t_{1}\ldots t_{i-1}$ and decode the next token $t_i$ autoregressively  as 

$$
    p(t_i \mid t_{1}\ldots t_{i-1}) = \Psi(t_{1}\ldots t_{i-1}),
$$
where $\Psi$ is the language model.
We provide two sources of rich instance-level information to the captioner: the instance-aware refined query output by the instance-segmentation model, and voxel features from 3D UNet backbone $F$. The refined query vector $Q_{r,o}$ corresponding to an object $o$ is first projected with a linear layer to generate the caption-aware query $Q_{c,o} \in \mathbb{R}^{D_{\mathrm{caption}}}$ for that object 
$$
    Q_{c,o} = \Phi_{\mathrm{query}}(Q_{r,o}),
$$
where $D_{\mathrm{caption}}$ is the the embedding dimension of the language model.
$Q_{c,o}$ contains localized information required to generate the caption for an object, such as its semantic class and its high-level features, and is used as the initial token to start the caption generation as

$$
    p(t_i \mid Q_{c,o},t_{1}\ldots t_{i-1}) = \Psi(Q_{c,o},t_{1}\ldots t_{i-1}).
$$

Additionally, we inject fine-grained information by using the captioner's cross-attention layers to attend to fine-grained contextual features from the 3D backbone $F$ corresponding to the object being captioned, after a linear projection to $D_{\mathrm{caption}}$. To reduce the computational complexity of the cross-attention operation, we aggregate features within precomputed segments \cite{felzenszwalb2004efficient} on the mesh surface, giving the context features $S_{o} \in \mathbb{R}^{n_{s,o} \times D_{\mathrm{caption}}}$, where $n_{s,o}$ is the number of segments in $I_o$, the predicted mask for object $o$. The cross-attention operation is $A=\mathrm{softmax}(qk^T/\sqrt{d})v$ where queries $q$ and values $k$ are the hidden states of $\Psi$, and the keys $k$ are $S_o$. The object and part captioners use context features $S_{\mathrm{obj},o}$ and $S_{\mathrm{part},o}$ respectively.

Importantly, to enable the object-level captioner to be informed by the part-level captioner, we make use of the last layer hidden states $[h_{\mathrm{part},1} \ldots h_{\mathrm{part},i}]$ of the part-level captioner $\Psi_{\mathrm{part}}$. We use a linear projection 

$$ 
H_{\mathrm{part},o} = \Phi_{\mathrm{hidden}}([h_{\mathrm{part},1}\ldots h_{\mathrm{part},i}]) 
$$

and then concatenate $H_{\mathrm{part},o}$ to the contextual features of the object captioner $S_{\mathrm{obj},o}$ for object $o$. The object-level caption for object $o$ is finally informed by the caption-aware query $Q_{c,o}$ as well as features $[H_{\mathrm{part},o}; S_{\mathrm{obj},o}]$.
Since different objects in a batch have different numbers of predicted segments, we zero-pad $S_{\mathrm{obj},o}$ to the maximum context length in the batch. 

The two caption models are trained with cross-entropy (CE) loss against the object- and part-level ground truth (GT) captions respectively, giving the loss 

$$
    \mathcal{L}_{\mathrm{caption}}=\mathcal{L}_{\mathrm{obj-caption}}+\mathcal{L}_{\mathrm{part-caption}}
$$

The outputs of this stage are the object-level caption $c_{\mathrm{obj},o}$, part-level caption $c_{\mathrm{part},o}$ and the corresponding language model hidden states $h_{\mathrm{obj}} = [h_{\mathrm{obj},1}\ldots h_{\mathrm{obj},i}]$ and $h_{\mathrm{part}}=[h_{\mathrm{part},1}\ldots h_{\mathrm{part},j}]$ for each object $o$.

\subsection{Semantic and Textual Consistency Losses}
While the object-level of detail is informed by the part-level of detail, supervising these two captioners with different GT captions may lead to inconsistencies in the predicted captions due to the different GT distributions of the two levels of detail. Hence, we use additional constraints in the form of consistency losses over the predicted captions. We encourage \textit{semantic consistency} for both captions to refer to the same object semantics, such as its fine-grained object class and material, and \textit{textual consistency} for the captions to be similar to each other.
Since consistency losses cannot be applied directly on the generated text due to the text tokenizer being non-differentiable, we instead use the language model hidden states $h_{\mathrm{obj}}$ and $h_{\mathrm{part}}$ from the last captioner layers as a proxy.

\paragraph{Semantic Consistency}
To promote both levels of captions to contain similar semantic information, we use a semantic consistency loss.
We do this by projecting the hidden states $h_{\mathrm{obj}}$ and $h_{\mathrm{part}}$ into a lower dimension to retain only meaningful semantic information over the whole caption. We then pass these sequences independently through object- and part-level caption projection transformers $\Phi_{\mathrm{obj}, \mathrm{sem}}$ and $\Phi_{\mathrm{part}, \mathrm{sem}}$ to obtain new embedded sequences, and 
 these aggregate into single feature vectors representing the overall semantics of the predicted captions $h_{\mathrm{obj},\mathrm{sem}} \in \mathbb{R}^{D_{\mathit{ff}}}$ and $h_{\mathrm{part},\mathrm{sem}} \in \mathbb{R}^{D_{\mathit{ff}}}$ respectively, where $D_{\mathit{ff}}$ is the feed-forward dimension of $\Phi$.
These are again independently projected into $N_{\mathrm{sem}}$ fine-grained latent classes using linear layers to get class probabilities $sem_{\mathrm{obj}}$ and $sem_{\mathrm{part}}$. The semantic consistency loss is a symmetric sum \cite{chen2021exploring} of cross-entropy (CE) losses

\begin{align*}
    \mathcal{L}_{\mathrm{semantic}} =  & \; CE(sem_{\mathrm{obj}}, SG(sem_{\mathrm{part}})) \; + \\ 
     & \; CE(sem_{\mathrm{part}}, SG({sem_{\mathrm{obj}}})),
\end{align*}
where $SG$ is the stop-gradient operator.

\paragraph{Textual Consistency}

The object- and part-level captions are expected to slightly overlap, since the salient features of an object are reflected in its object-level caption. We encourage this similarity through the similarity of the whole sequences of object-level and part-level captions. Similar to the semantic consistency loss, we use projectors $\Phi_{\mathrm{obj},\mathrm{text}}$ and $\Phi_{\mathrm{part},\mathrm{text}}$ to project the language model hidden states into $h_{\mathrm{obj},\mathrm{text}}$ and $h_{\mathrm{part},\mathrm{text}}$, and aggregate over the sequence of tokens to get single feature vectors $\overline{h}_{\mathrm{obj},\mathrm{text}}$ and $\overline{h}_{\mathrm{obj},\mathrm{part}}$. The text-consistency loss is the distance $d(.,.)$ between these two feature vectors 

$$
    \mathcal{L}_{\mathrm{textual}} = d(\overline{h}_{\mathrm{obj},\mathrm{text}}, \overline{h}_{\mathrm{part},\mathrm{text}}).
$$

The final loss used to optimize \OURS{} is a weighted sum of the individual losses
$$
    \mathcal{L} = w_1\mathcal{L}_{\mathrm{caption}} + w_2\mathcal{L}_{\mathrm{semantic}} + w_3\mathcal{L}_{\mathrm{textual}}
$$

\paragraph{Implementation Details}
Predicted instances are matched with GT instances using Hungarian matching \cite{kuhn1955hungarian}, and the predicted captions are supervised by the captions of the matched instances. 
We train both the instance segmentation and captioning stages with a batch size of 6 scenes. In the captioning stage, one batch that is input to the captioner contains all objects from scenes in the batch. We use a cosine annealing learning rate schedule starting with a learning rate of $0.005$, and an AdamW optimizer. Training takes about 1 day. All experiments are carried out on an Nvidia A6000 GPU.

\section{\DatasetName{}}
\label{sec:dataset}

Existing datasets for captioning objects in 3D indoor scenes  contain captions at a single level of detail, that combines both object properties such as color and texture, as well as spatial and functional relations to other objects in the scene. To enable the task of captioning at both the object- and part-levels of detail, we created the \DatasetName{} based on the ScanNet++~\cite{yeshwanthliu2023scannetpp} dataset of 3D indoor scans. The data is generated in a scalable and fully automated manner by leveraging a VLM that takes an  RGB image crop of an object and a text prompt as input, and outputs a description of the object. The objects to be captioned are identified using the ground truth semantic instance annotations provided by ScanNet++. We prompt the VLM with text prompts to generate specific information about an object's color, texture, material and similar properties, and leverage multi-view reasoning to obtain reliable final caption generation for ScanNet++. A comparison of the \DatasetName{} with existing object-caption and other 3D object-text datasets is shown in Tab. \ref{tab:dataset_comparison}.

\begin{table}[t]
    \centering

    \begin{tabular}{lcccc}
        \toprule
        Dataset &  \#Desc & \#Cls & \#Objs & LoD \\ %
        \midrule
        
        NR3D \cite{achlioptas2020referit_3d} &  41k & 76 & 5.8k & s + o \\ %
        Multi3DRefer \cite{zhang2023multi3drefer}  & 55k & 265 & 10.5k & s + o  \\ %
        ScanQA \cite{azuma2022scanqa} & 35k & 370 & 9.5k & s + o  \\
        Scan2Cap \cite{chen2021scan2cap} & 46k & 265 & 9.9k & s + o  \\
        Ours & \textbf{\DatasetNumTotalCaptions{}} & \textbf{2k} & \textbf{34.7k} & o + p  \\ %
        \bottomrule
    \end{tabular}
    \caption{Comparison of the \DatasetName{} with existing datasets that describe objects in 3D indoor scenes. We compare the number of descriptions (\#Desc), number of unique object classes described (\#Cls), number of objects described (\#Objs) and level of detail (LoD, s = scene, o = object, p = part). Our descriptions have greater object detail and introduce part-level details, while scaling up the number of descriptions, and semantic classes and objects described.}
    \label{tab:dataset_comparison}
    \vspace{-0.35cm}
\end{table}

\subsection{Object Level of Detail Captioning}
We utilize the ground truth 3D instance and semantic annotations in the ScanNet++ dataset to generate the \DatasetName{}. We project the 3D annotations onto the high-resolution DSLR images that are registered with the 3D scan, to obtain a corresponding 2D bounding box of each object in each DSLR image. We use this 2D mask to crop a rectangular region containing the object from the image, after expanding the bounding box dimensions by $10\%$ .
To maximize the RGB information that is input to the VLM, we pick the top 3 images that cover atleast $10\%$ the mesh vertices corresponding to the object mask. The VLM generates $k=5$ descriptions for each view of the object, giving outputs $c_{i,j}$ from the $i^{\mathrm{th}}$ view. We use an instruction-tuned large language model (LLM) to obtain the $j^{th}$ object-level caption $\hat{c}_{\mathrm{obj},j}$ by combining the $j^{th}$ caption from each view.

\begin{figure}
\centering
\includegraphics[width=\linewidth]{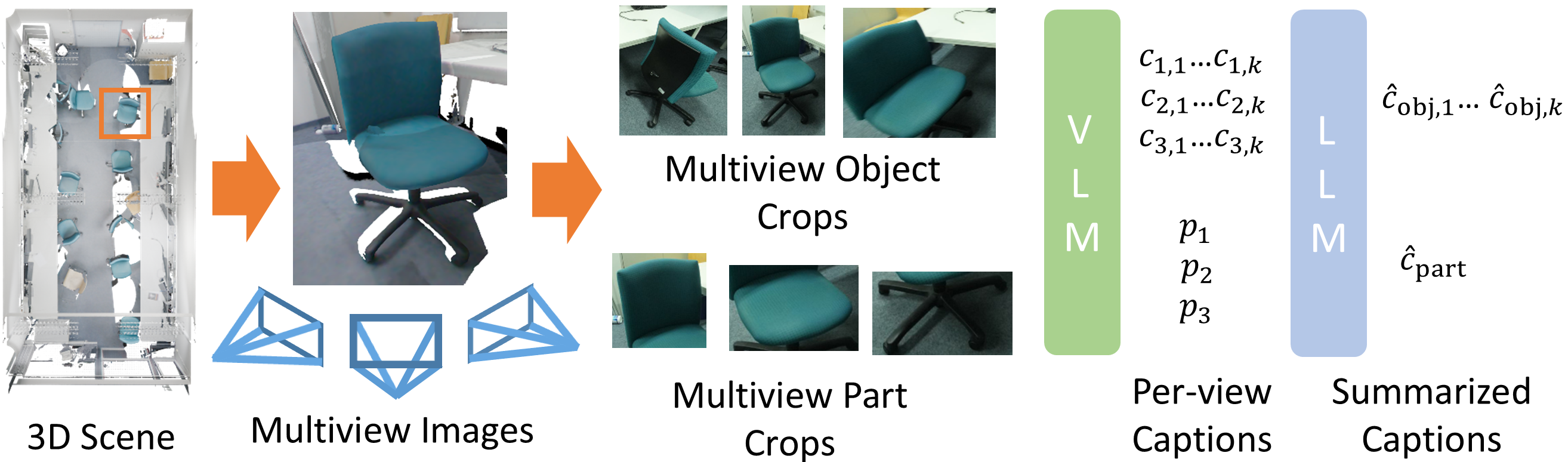}
\caption{Generation of \DatasetName{}. We use the ground truth 3D semantics in ScanNet++, project them onto multiview DSLR images and obtain descriptions of the image crops $c_{i,j}$ using a VLM. For parts crops, we use pseudo-ground truth from MaskClustering. Finally, we summarize the captions from different views using an LLM.}
\label{fig:multiview-data-generation}
\end{figure}

\subsection{Part Level of Detail Captioning}
We produce pseudo-ground truth 3D part segmentation data using a modified version of MaskClustering \cite{yan2024maskclustering} that favours smaller segments over larger ones, along with SAM \cite{kirillov2023segment}. SAM is used to generate initial 2D segments on the DSLR images. MaskClustering combines these 2D segments to produce coherent 3D part segments. We associate these 3D parts with the objects in ScanNet++ based on an intersection-over-union (IoU) threshold. The rest of the part captioning process largely follows that of object captioning. In addition to a cropped image of the part, we generate a second caption based on a context image that includes the whole object, with the part indicated by a red bounding box. This improves caption quality when the crop of the part is small. We generate one caption $p_i$ in each view, for each part, and then combine the descriptions of all the parts into a single part-level caption $\hat{c}_{\mathrm{part}}$ for each object using the LLM.
While the pseudo-masks at the part-level obtained from MaskClustering can detect parts of long-tail and small objects in the ScanNet++ data, these masks are not as reliable as the manually annotated object GT masks in ScanNet++. Hence, we do not use the pseudo-masks to directly predict part segments during captioning, and instead aggregate the captions of different parts into a single caption containing part-level details.
\vspace{-0.25cm}
\paragraph{Implementation Details}
We use Llava 1.6 \cite{liu2024llavanext} (7B parameters) with a Mistral backbone as the VLM, and instruction-tuned Llama 3.1 \cite{dubey2024llama} (8B parameters) for the LLM. Since the generated captions in the \DatasetName{} are relatively long and detailed, we simplify the captions for training to a shorter length using an LLM.

\vspace{-0.25cm}
\section{Experiments}
\label{sec:results}

\begin{figure*}
\centering
\includegraphics[trim=0 3.2cm 0 0, clip, width=0.95\textwidth]{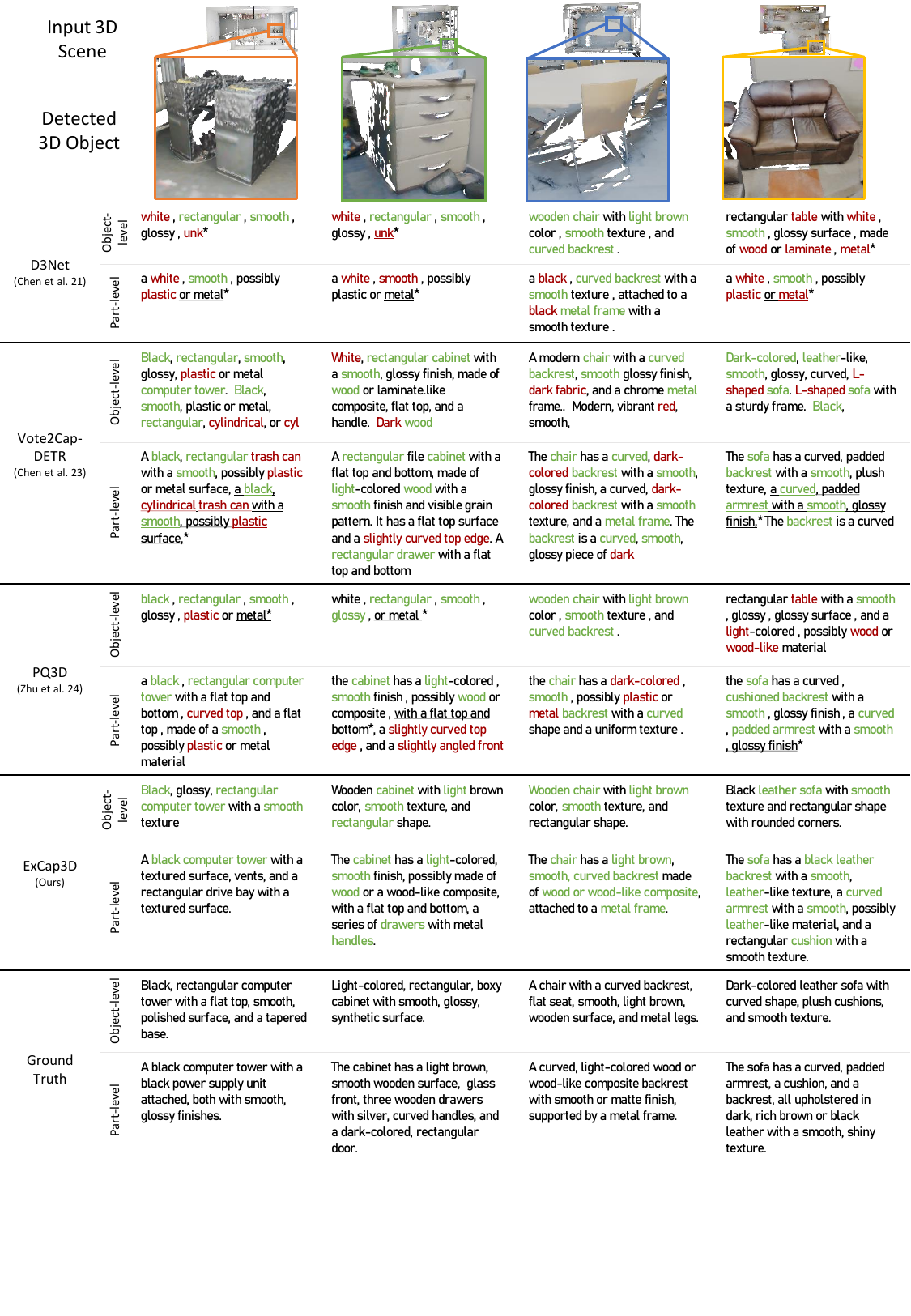}
\caption{Qualitative evaluation on ScanNet++~\cite{yeshwanthliu2023scannetpp} scenes, in comparison with D3Net~\cite{chen2022d3net}, Vote2CAP-DETR~\cite{chen2023vote2cap}, and PQ3D~\cite{zhu2024unifyingpq3d}. Predicted text from methods are denoted by color to indicate \textcolor{ForestGreen}{correct} and \textcolor{BrickRed}{incorrect} generated phrases compared to the ground truth. Our method produces more consistent and detailed captions at both the object- and part-levels of detail. * indicates truncated output where the underlined phrase was predicted repeatedly. 
}
\label{fig:qual_results}
\end{figure*}

We evaluate caption generation on the 84 instance class categories of the ScanNet++ benchmark~\cite{yeshwanthliu2023scannetpp}.  We use the train/val/test splits of the semantic tasks of ScanNet++.

\subsection{Evaluation Metrics}
We adopt the commonly used CIDEr \cite{santos2021cider}, ROUGE \cite{lin2004rouge}, and METEOR \cite{denkowski2014meteor} caption evaluation metrics used by other 3D captioning methods \cite{chen2021scan2cap, chen2023vote2cap, zhu2024unifyingpq3d, zhu20233d, huang2024chat}. These metrics are averaged over every GT instance, with a weight of 1 if the IoU of the object's predicted bounding box with the ground truth bounding box is greater than $0.5$, and 0 otherwise. Metrics are reported separately for the object- and part-level details.

\subsection{Baselines}
We compare our method against three state-of-the-art captioning methods. D3Net~\cite{chen2022d3net} uses a PointGroup sparse convolutional backbone with embedded object relations, and a GRU head for text generation. Vote2Cap-DETR~\cite{chen2023vote2cap} tokenizes the input point cloud into a fixed number of point tokens, encoded with a transformer, and then decoded into captions using a GPT2 \cite{radford2019language} language model. PQ3D~\cite{zhu2024unifyingpq3d} uses promptable queries to encode regions of interest, and aggregates 3D features to generate captions with a T5-Small \cite{raffel2020exploring} language model. 

For all baselines, as well as our method, we pretrain the respective detection or segmentation backbones on  ScanNet~\cite{dai2017scannet}. We then train on ScanNet++~\cite{yeshwanthliu2023scannetpp} for bounding box detection (Vote2Cap and D3Net) or instance segmentation (PQ3D and \OURS{}). 
For voxel-based methods (D3Net, PQ3D and \OURS{}), we use a voxel size of $2\mathrm{cm}$. We use a graph-cut based oversegmentation \cite{felzenszwalb2004efficient} to precompute segments for PQ3D and \OURS{}. 
Since the baselines were designed for single level of detail captioning, we train them separately on the object- and part-level captions, while \OURS{} is trained jointly on both levels of detail. We use only 3D scan inputs for all methods.

\begin{table}[h]
    \centering
    \begin{tabular}{lccc}
        \toprule
        & \multicolumn{3}{c}{\textbf{Object-level Details}} \\
        \cmidrule(lr){2-4}
        Method & CIDEr & ROUGE & METEOR  \\
        \midrule
        D3Net \cite{chen2022d3net} & 6.7	&5.4	&6.7		\\
        Vote2Cap-DETR \cite{chen2023vote2cap} & 13.3&	\underline{12.9}&	\underline{17.2}	\\
        PQ3D \cite{zhu2024unifyingpq3d} & \underline{27.9}	&11.6	&12.5	\\
        ExCap3D \textit{(Ours)} & \textbf{32.7}&	\textbf{16.6}&	\textbf{17.9} \\
    \end{tabular}
    \begin{tabular}{lccc}
        \toprule
        & \multicolumn{3}{c}{\textbf{Part-level Details}} \\
        \cmidrule(lr){2-4}
        Method & CIDEr & ROUGE & METEOR \\
        \midrule
        D3Net \cite{chen2022d3net} & 10.5	&7.9	&7.9	 \\
        Vote2Cap-DETR \cite{chen2023vote2cap} &	13.3&	\underline{20.7}&	\textbf{22.7} \\
        PQ3D \cite{zhu2024unifyingpq3d} & \underline{14.4}	&16.3	&15.6 \\
        ExCap3D \textit{(Ours)} & \textbf{32.3}&	\textbf{21.7}&	\underline{20.8} \\
        \bottomrule
    \end{tabular}
    \caption{Comparison of object-level and part-level captioning performance between \OURS{} and baselines. The second-best result is \underline{underlined}. \OURS{} improves CIDEr scores significantly through joint captioning and consistency losses. All metrics are evaluated at an IoU threshold of 0.5}
    \label{tab:captioning_results}
\end{table}

\subsection{Comparison to State of the Art}
We compare with state-of-the-art methods D3Net~\cite{chen2022d3net}, Vote2Cap-DETR~\cite{chen2023vote2cap}, and PQ3D~\cite{zhu2024unifyingpq3d} in Tab.~\ref{tab:captioning_results}. \OURS{} outperforms all baselines on the CIDEr score, which aligns the most closely with human perception, improving upon the highest performing baseline PQ3D by a margin of \ObjectCiderImprovementPercentage\% and \PartCiderImprovementPercentage\% for object- and part-level details respectively. We also outperform the baselines on the ROUGE and METEOR scores of object-level details, and ROUGE scores of part-level details.
Our joint prediction of captions at different levels of detail enables more robust, accurate descriptions.

Qualitative comparisons are shown in Fig. \ref{fig:qual_results}. \OURS{} produces complete and detailed descriptions at both the object- and part-levels of detail, while the baseline methods often omit details in both levels. D3Net uses a limited GLOVE embedding of input text that primarily measures similarity of words along with limited per-instance features, and hence cannot produce detailed output descriptions. 
Vote2Cap-DETR and PQ3D interpret 3D scenes more coarsely, which can result in hallucination of incorrect fine-grained details.

\subsection{Ablation Studies}
We ablate the different components of our method and show their contribution towards the final captioning model and show the results in Tab.~\ref{tab:ablation}. As a baseline, we use two independent captioning heads and train with only $\mathcal{L}_{\mathrm{caption}}$. Each component of our method is added separately to the baseline to show its efficacy. 

\vspace{-0.3cm}

\paragraph{Enforcing multilevel consistency improves caption quality}
Semantic consistency primarily improves part-level details while textual consistency primarily improves object-level details as shown in Tab.~\ref{tab:ablation}, which demonstrates the complementary nature of   these two losses. As the object-level captions typically contain a  stronger notion of global object semantics compared to the part-level caption, the latter improves in quality with  semantic consistency. The object-level captions also receive fine-grained cues through textual similarity from the part-level details which helps to refine the global description of the object.

\vspace{-0.3cm}

\paragraph{Part-object captioner information sharing improves performance}
Adding part $\rightarrow$ object information sharing by conditioning object captions on the part-level of detail improves both levels of captions as shown in Tab.~\ref{tab:ablation}. This is because object details are improved when they are informed by part captions which contain aggregated details that can produce a better global description of the object. At the same time, the part-level of detail improves as it is guided to produce descriptions of parts that can be useful for object-level captioning and are consistent as a whole. 

\vspace{-0.3cm}

\paragraph{Consistency and information sharing are complementary to each other}
The semantic and textual consistency losses act independently of the part $\rightarrow$ object information sharing mechanism, and hence our full model improves significantly over the individual components, particularly in the part-level of detail as shown in Tab.~\ref{tab:ablation}. Intuitively, this is because the consistency losses model only two kinds of consistency in a lower-dimension embedding space, while the sharing of language model hidden states enables the object-level captioner to directly condition on the internal high-dimension representation of the part-level captions which contains fine-grained text and 3D object features.

\begin{table}[t]
    \centering
    \begin{tabular}{lcc}
        \toprule
        Method & Object-Level & Part-level \\
        \midrule
        Separate Models & 29.8 & 18.7 \\
        w/ Semantic Consistency & 30.2  & 24.8 \\
        w/ Textual Consistency & 32.2 & 19.6 \\
        w/ Part $\rightarrow$ Obj. Inf. Sharing & \textbf{34.8} & \underline{25.4} \\
        Full Model (Ours) & \underline{32.7} & \textbf{32.3} \\
        \bottomrule
    \end{tabular}
    
    \caption{Ablation study of the different components of our method. Performance is reported using CIDEr for object-level and part-level captions. Each ablation is applied independently to the baseline with separate models.}
    \label{tab:ablation}
\end{table}

\vspace{-0.3cm}

\paragraph{Limitations}
While our approach generates expressive captioning for 3D scans, various limitations remain.
We use two captioning models, with one conditioned on the hidden states of the other through cross-attention. This could limit the amount of information shared between the captioners.
Our method relies on a sparse convolutional backbone \cite{choy20194minkowski} which is limited to a resolution of about $2$cm. This could restrict captioning performance on small or thin objects.

\section{Conclusion}
\label{sec:conclusion}
We introduced \OURS{}, a method for expressive dense captioning of 3D scenes at multiple object- and part-levels of detail. We showed that describing 3D objects as the sum of their parts while maintaining consistency between the multiple levels of descriptions leads to more accurate descriptions aligned with human perception. We hope that our method and the accompanying \DatasetName{} open up new possibilities for a holistic, expressive understanding of 3D scans.

\section*{Acknowledgments}
This project was supported by the ERC Starting Grant SpatialSem (101076253), and the German Research Foundation (DFG) Grant ``Learning How to Interact with Scenes through Part-Based Understanding."

{
    \small
    \bibliographystyle{ieeenat_fullname}
    \bibliography{main}
}

\clearpage
\setcounter{page}{1}
\maketitlesupplementary

\begin{figure*}
\centering
\includegraphics[trim=0 10cm 0 0, clip, width=\textwidth]{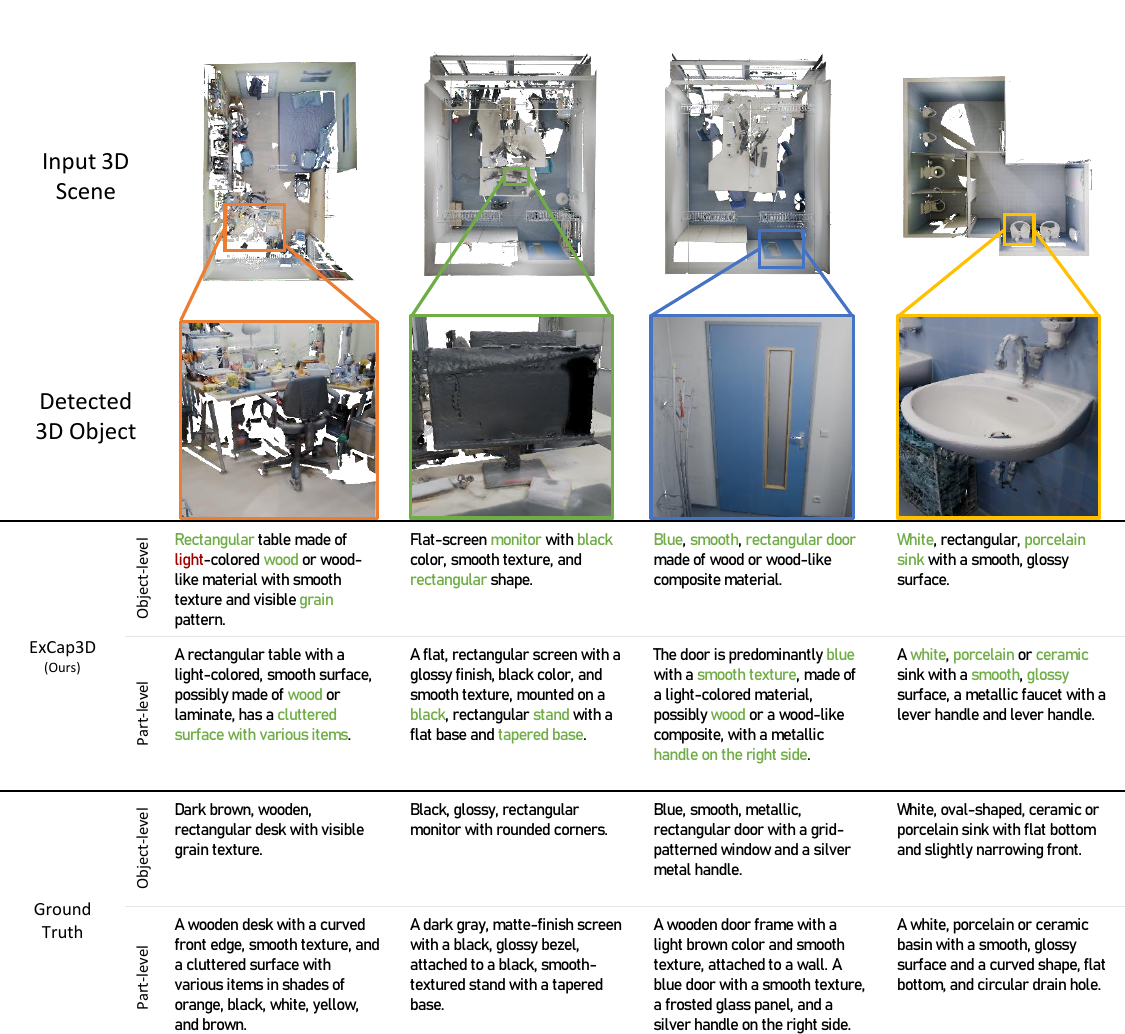}
\caption{Object- and part-level captions predicted by \OURS{} on a diverse set of semantic classes and indoor scenes. \OURS{} accurately describes both part-level properties of objects such as material and texture, as well as object-level properties.
}
\label{fig:suppl_qual_results}
\end{figure*}

\section{Additional Results}
\paragraph{Modeling objects as a sum of parts improves caption quality}
\OURS{} models an object as a sum of its parts by conditioning the object captioner on the outputs of the part captioner. Alternately, we can model top-down with the parts being components of a whole, by first generating object descriptions and conditioning the part captioner on the corresponding hidden states. Results of this reversed information flow are shown in Tab.~\ref{tab:direction-sharing}. Other consistency losses are kept the same in both experiments. Part-level details do not benefit from object $\rightarrow$ part information sharing as object-level captions do not contain fine-grained part information, while part captions contain information that is useful for describing the whole object.

\begin{table}[h]
    \centering
    
    \begin{tabular}{lcc}
        \toprule
        Method & Object-Level & Part-level \\
        \midrule
        Object $\rightarrow$ Part Info. Sharing & \textbf{32.8} & 15.6 \\
        Part $\rightarrow$ Object Info. Sharing & 32.7 & \textbf{32.3} \\
        \bottomrule
    \end{tabular}
    \caption{Comparison of different directions of information sharing between the object and part captioners. Performance is reported in CIDEr@0.5.}
    \label{tab:direction-sharing}
\end{table}

\paragraph{Fine-grained context features improve part captioning}
\OURS{} uses segment-level object context features for captioning. This is required particularly for describing low-level part details of smaller regions of objects, in addition to the caption-aware query features $Q_{c,o}$ which contain coarse information about the whole object. Results are shown in Tab. \ref{tab:context-feats}. Object-level details are sufficiently captured by the query features, while part details benefit from the fine-grained per-segment context features.

\begin{table}[h]
    \centering
    
    \begin{tabular}{lcc}
        \toprule
        Method & Object-Level & Part-level \\
        \midrule
        w/o context features & \textbf{33.7} & 27.0 \\
        w/ context features & 32.7 & \textbf{32.3} \\
        \bottomrule
    \end{tabular}
    \caption{Comparison of our model with and without object context features. Performance is reported in CIDEr@0.5.}
    \label{tab:context-feats}
\end{table}

\paragraph{End-to-end learned captions improve upon separate segmentation and VLM models} \OURS{} combines the tasks of instance segmentation and object captioning. Instead, we can first predict instance masks using Mask3D and describe them with the VLM by projecting them onto the multiview DSLR images. However, these instance predictions are not as accurate as the GT instances and can limit the captioning performance of the VLM which expects precise object crops. Alternately, we can render the 3D mesh at the DSLR image camera poses and use these as input to the VLM. Since VLMs are primarily trained on natural images, this is expected to give low quality captions. In both cases, the VLM may produce slight inconsistencies at the different levels of detail. Results are shown in Tab. \ref{tab:predinst-vlm}. \OURS{}'s end-to-end and learned captioning approach outperforms the VLM applied on DSLR images or rendered images of meshes.

\begin{table}[h]
    \centering
    
    \begin{tabular}{lcc}
        \toprule
        Method & Obj & Part \\
        \midrule
        Pred. Inst. + VLM on DSLR images & 18.6 & 11.4 \\
        Pred. Inst. + VLM on rendered mesh  & 21.4 & 15.5 \\
        End-to-end captioning & \textbf{32.7} & \textbf{32.3} \\
        \bottomrule
    \end{tabular}
    \caption{Comparison of our model with the VLM applied on predicted 3D instances, projected to multiview DSLR images and multiview 3D mesh renders. Performance is reported in CIDEr@0.5 at object- (Obj) and part-levels (Part).}
    \label{tab:predinst-vlm}
\end{table}

 \paragraph{Qualitative results}
 We show additional qualitative results from \OURS{} in Fig. \ref{fig:suppl_qual_results}. \OURS{} makes use of fine-grained 3D features on predicted instances to predict expressive object and part properties such as material, texture and appearance.

\section{Implementation Details of \OURS{}}

\paragraph{Instance segmentation}
We use the default configuration of Mask3D for instance segmentation training, with 100 instance queries and an embedding dimension of 128. Since ScanNet++ contains several large and complex scenes, we randomly sample $300k$ points for training on such scenes, while evaluation is carried out on all points. Training takes about 4 days  on an Nvidia A6000. The instance segmentation model has an AP50 score of $0.38$ on the ScanNet++ validation set with $84$ instance classes. 

\paragraph{Multilevel captioning}
The instance segmentation model is kept frozen during caption training. 
During training we sample from the different captions for an object, and during inference we use a single fixed caption as the GT. Since all the objects in a training batch may not have captions after being filtered out by visibility constraints, $\mathcal{L}_{\mathrm{caption}}$ is applied only on the objects that have captions. During inference, captions are produced for all $100$ input queries and matched with GT objects that have captions.
For the object- and part-level captioners $\Psi_{\mathrm{obj}}$ and $\Psi_{\mathrm{part}}$ we use a language model with the GPT2 architecture trained from scratch, with an embedding size of 128, 1 layer and 4 attention heads. 

For the consistency losses, caption projection models $\Phi_{\mathrm{obj}}$ and $\Phi_{\mathrm{part}}$ we use transformer encoders with 2 layers each, an embedding dimension of 16, feedforward dimension of 128 and 2 attention heads. The loss weights are determined as $w_1=1, w_2=w_3=0.1$ empirically.
 During inference we use beam search with 5 beams, and pick hidden states from the beams corresponding to the finally predicted caption.

 \section{Implementation Details of Data Generation}

\paragraph{Part segmentation pseudo-mask generation}
To generate 2D masks with SAM \cite{kirillov2023segment}, we prompt it with points on the DSLR images. These points are sampled from the corresponding 3D precomputed segments on the mesh \cite{felzenszwalb2004efficient} to encourage consistency between masks from different views. Then we use MaskClustering \cite{yan2024maskclustering} to backproject and combine the 2D masks on the 3D mesh and keep the smaller mask when two masks overlap. We subsample the densely captured DSLR images by a factor of $5$ which resulted in sufficient multiview consensus. 

\paragraph{Caption dataset generation}
Since ScanNet++ contains a dense capture trajectory of DSLR images, we subsample every $10^{\mathrm{th}}$ DSLR frame from the train and val splits. 
In addition to filtering the DSLR images for visibility of the object, we filter out bounding boxes with a dimension of less than $50$ pixels to avoid very small inputs to the VLM. 
To avoid the VLM describing the distortion in the fisheye DSLR capture of ScanNet++ (e.g., \textit{this is a distorted image of a table}) and match the training distribution of the VLM, we undistort the images using the provided distortion parameters before using them as input to the VLM.

\paragraph{VLM prompts for caption generation}
We prompt the VLM with multiview crops of each object, along with its semantic class name and 3D dimensions in meters. The VLM is prompted to generate the object's shape, structure, color and texture using only information available in the image, without adding any common-sense information. For object part crops we provide the name of the object in the VLM's input. 

\paragraph{LLM prompts for multiview and part caption aggregation}
To aggregate over captions generated in multiple views, we prompt the LLM with a list of these captions and to output the most likely and uniquely identifying information of the object/part, and remove any commonsense information about it. For object parts, we first aggregate over multiple views, and then over all the parts belonging to an object to get a single part-level caption for the object.
We also experimented with aggregation of the object-level captions within each view and then over all views, and found the resulting captions to be of lower quality.

\section{Captioning Metrics}
We provide a brief overview of the different captioning metrics used.
\paragraph{ROUGE}
ROUGE is based on BLEU \cite{papineni2002bleu}, which measures the overlap of $n$-grams in the predicted and GT captions. ROUGE adds additional precision and recall terms to BLEU. We use the F1 score of the ROUGE-L variant that measures matches of longest common subsequences, even if the matching words are not contiguous. ROUGE focuses on caption recall, and hence rewards long captions that capture as much content of the GT as possible.
ROUGE ranges from $0$-$1$.

\paragraph{METEOR}
METEOR \cite{denkowski2014meteor} adds the notion of semantic matching of words between candidate predictions and GT based on a fixed database \cite{miller1995wordnet}, and computes matches between chunks of words that need not be contiguous. However, it has a larger weight for $n$-gram recall compared to precision.
METEOR ranges from $0$-$1$.

\paragraph{CIDEr}
CIDEr \cite{santos2021cider} is a widely used metric for evaluating caption predictions against GT captions in the 2D and 3D domains. It has been shown to correlate highly with human judgement. The initial version of CIDEr \cite{vedantam2015cider} improved upon ROUGE and METEOR by incorporating a TF-IDF term over the whole GT corpus, and hence downweights $n$-grams that occur frequently in the GT. This increases the weight of terms that occur rarely in the GT, and occur in the prediction. In our case, these rare terms are the unique and identifying features of the objects. We use CIDEr-R \cite{santos2021cider}, simply written as CIDEr, which adds additional length and repetition penalties on the $n$-grams. The final score is averaged over multiple values of $n$. CIDEr score ranges from $0$-$10$, due to a normalizing factor of $10$ to make it comparable with other metrics.

\paragraph{Display convention} We report results of all three metrics after multiplying by $100$, as done in other 3D dense captioning works \cite{zhu2024unifyingpq3d, chen2021scan2cap, chen2023vote2cap}.

\end{document}